\documentclass{article}
\usepackage{spconf,amsmath,amsfonts,graphicx}
\usepackage[colorlinks=true]{hyperref}
\usepackage{enumitem}
\usepackage{arydshln}
\usepackage{bm}
\usepackage{multirow}
\usepackage{color}

\setlist{nosep, leftmargin=14pt}

\usepackage{mwe} 

\title{Towards Spatial Transcriptomics-guided \\ Pathological Image Recognition with Batch-Agnostic Encoder}
\name{Kazuya Nishimura$^{\dagger}$, Ryoma Bise$^{\ddagger}$, Yasuhiro Kojima$^{\dagger}$}
\address{$^{\dagger}$National Cancer Center Japan~ $^{\ddagger}$Kyushu University}
\begin{document}
%
\maketitle
\begin{abstract}
Spatial transcriptomics (ST) is a novel technique that simultaneously captures pathological images and gene expression profiling with spatial coordinates. Since ST is closely related to pathological features such as disease subtypes, it may be valuable to augment image representation with pathological information. However, there are no attempts to leverage ST for image recognition ({\it i.e,} patch-level classification of subtypes of pathological image.). One of the big challenges is significant batch effects in spatial transcriptomics that make it difficult to extract pathological features of images from ST.
In this paper, we propose a batch-agnostic contrastive learning framework that can extract consistent signals from gene expression of ST in multiple patients. 
To extract consistent signals from ST, we utilize the batch-agnostic gene encoder that is trained in a variational inference manner.
Experiments demonstrated the effectiveness of our framework on a publicly available dataset.
Code is publicly available at \url{https://github.com/naivete5656/TPIRBAE}

\end{abstract}
\begin{keywords}
Pathological image, spatial transcriptomics, contrastive learning, multi-modal learning
\end{keywords}
\section{Introduction}
\label{sec:intro}
\begin{figure}[t]
    \centering
    \includegraphics[width=0.9\linewidth]{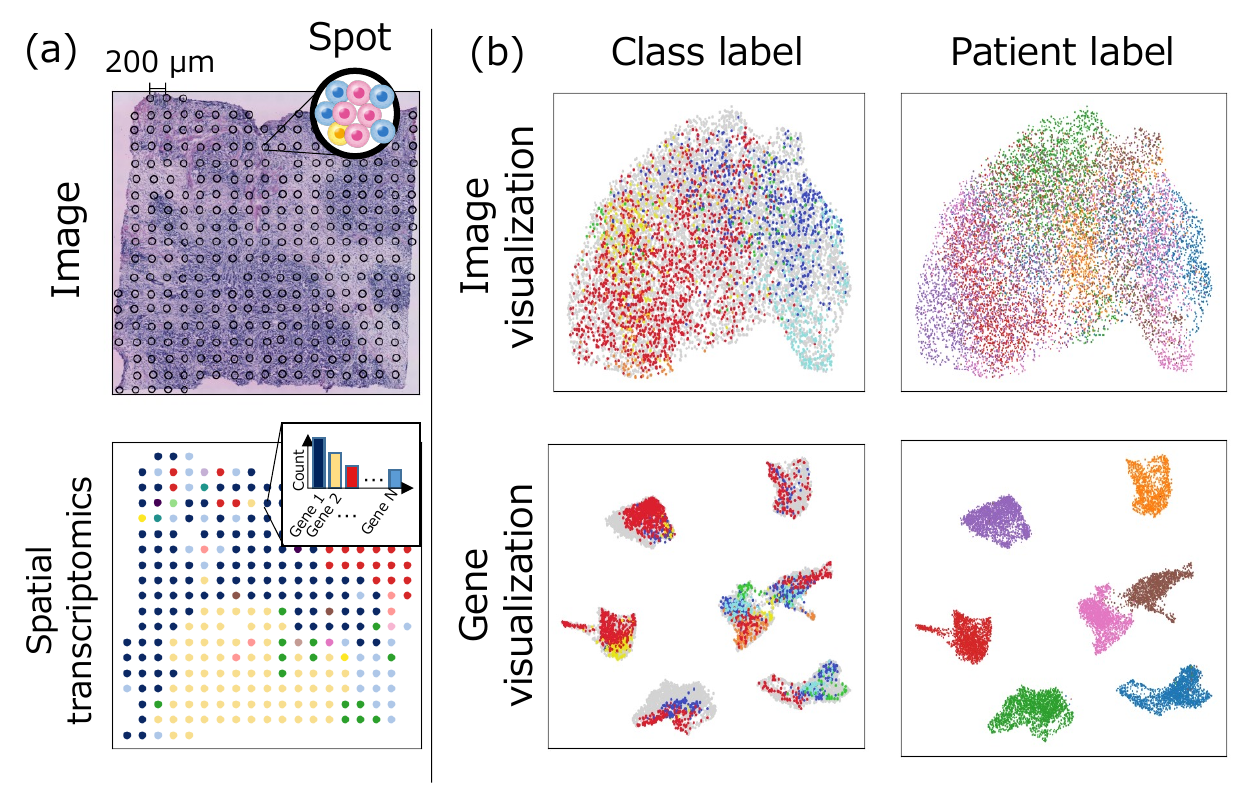}
    \caption{(a) A pair of a whole slide image and spatial transcriptomics. Gene expression is captured on spots, which contain the expression levels of tens of thousands of genes and with spatial coordinates. (b) Visualization of data distribution using UMAP projection on her2st dataset \cite{andersson2020spatial}. Left and right are colorized by subtype class labels for image and patient labels.} 
    \label{fig:intro}
\end{figure}

Spatial transcriptomics (ST) technology \cite{marx2021method} now enables gene expression profiling with spatial resolution accompanying pathological images. As shown in Fig. \ref{fig:intro} (a), the gene expression is observed on the spot with spatial coordinates, and the spot contains a count of tens of thousands of genes.
Gene expression indicates various cellular activities such as cancer subtypes \cite{khan2023deepgene, weng2023epigenetically}, survival rates \cite{hendam2023effects}, and toxicity \cite{alexander2018developments}, and also contains rich information for morphological features.
Consequently, integrating gene expression with images can potentially augment image recognition.

While a method using slide-level transcriptomics for slide-level image representation has been proposed \cite{jaume2024transcriptomics}, methods that can leverage spatial transcriptomics for patch-level representation have not been proposed.
This is largely due to the sensitivity of the ST techniques to noise and experiment batches compared to the slide-level data. Notably, there are large batch effects, which are data gaps due to non-biological factors such as experimental conditions. In fact, as shown in Fig. \ref{fig:intro} (b), the distribution of gene expression is separated by the patients due to the batch effects, and the trend is different from the distribution of the image.

In this paper, we aim to use contrastive learning with images and ST for image recognition ({\it i.e.,} subtype classification).
While recent studies have introduced contrastive learning methods to estimate gene expression from images by integrating these data \cite{xie2024spatially}, these methods overlook the problem of the batch effects and have not assessed how these effects influence image recognition performance. Using the raw data without correcting batch effect can lead to inconsistent signal extraction across experimental batches. 
To our knowledge, this is the first attempt to leverage ST for image recognition.

For the batch-agnostic encoder, we introduce a single-cell Variational Autoencoder (scVI \cite{lopez2018deep}), which is widely used for batch effect correction of single-cell data. By leveraging the batch-agnostic encoder for contrastive learning, we achieve batch-agnostic contrastive learning for image recognition.
In our experiments, we validated the effectiveness of the proposed framework and compared the effect of the proposed framework with the conventional methods. 
Our results demonstrate the value of batch effect correction and highlight key challenges in applying gene expression data for image recognition.

Our contributions are as follows:
\begin{itemize}
    \item We tackle a novel setting that uses images and spatial transcriptomics (ST) for image recognition. This task is challenging because we should capture consistent signals from the gene expression of ST among patients.
    \item We propose a novel contrastive learning framework for image and ST by repurposing an encoder trained by a variational inference method to extract consistent signals from multiple patients. 
    \item We evaluate the effectiveness of our framework using a public dataset. The results demonstrated that our approach outperforms previous contrastive learning-based methods, highlighting the effectiveness of both the batch correction and the class-guided learning method.    
\end{itemize}

\vspace{-2mm}
\section{Related work}

{\bf Multi-modal learning with image and transcriptomics.} 
James {\it et al.} \cite{jaume2024transcriptomics} have proposed transcriptomics-guided slide-level representation learning using slide-level transcriptomics ({\it i.e.,} bulk). Unlike slide-level transcriptomics, ST contains spatial coordinates and rich signals for cellular activity. 

Gene expression estimation from pathological images is one of the main streams for multi-modal learning with imaging and ST \cite{he2020integrating, pang2021leveraging, xie2024spatially, chung2024accurate}. 
Convolutional network \cite{he2020integrating} and transformer \cite{radford2021learning, chung2024accurate} have been introduced for gene expression estimation.
Contrastive learning-based methods also have been proposed \cite{xie2024spatially} to estimate gene expression. 
However, these methods do not consider the batch effect and could not extract consistent semantic signals from multiple patients.  

In contrast, we focus on leveraging ST for image recognition by addressing the batch effect. To our knowledge, this is the first attempt to utilize ST for the classification of images.

\noindent
{\bf Batch correction.}
Due to the technical complexity of single-cell transcriptomics and ST, there tends to be a more significant gap between experimental batches than traditional slide-level transcriptomics due to technical factors.
To align the distribution gap made by the batches, batch correction methods have been proposed \cite{korsunsky2019fast, lopez2018deep, xu2021probabilistic}. 
Variational inference of latent representations has been introduced to this field \cite{lopez2018deep} (scVI). Moreover, conditional VAE, which can reflect annotation information, has also been proposed \cite{xu2021probabilistic} to utilize annotation information for batch correction (scANVI).

In this paper, we utilized the encoder scVI and scANVI for batch-agnostic contrastive learning of image and ST, and we demonstrated the effectiveness of the framework.

\section{Batch-agnostic contrastive Learning}
\label{sec:foot}

The motivation of our method is to train feature extractor for image recognition ({\it i.e.,} classification of subtypes from image) with pairs of images and spatial transcriptomics. Since the raw gene expression has a large gap among experimental batches, we propose a novel contrastive framework by introducing a batch-agnostic encoder for ST.

\begin{figure}[t]
    \centering
    \includegraphics[width=0.9\linewidth]{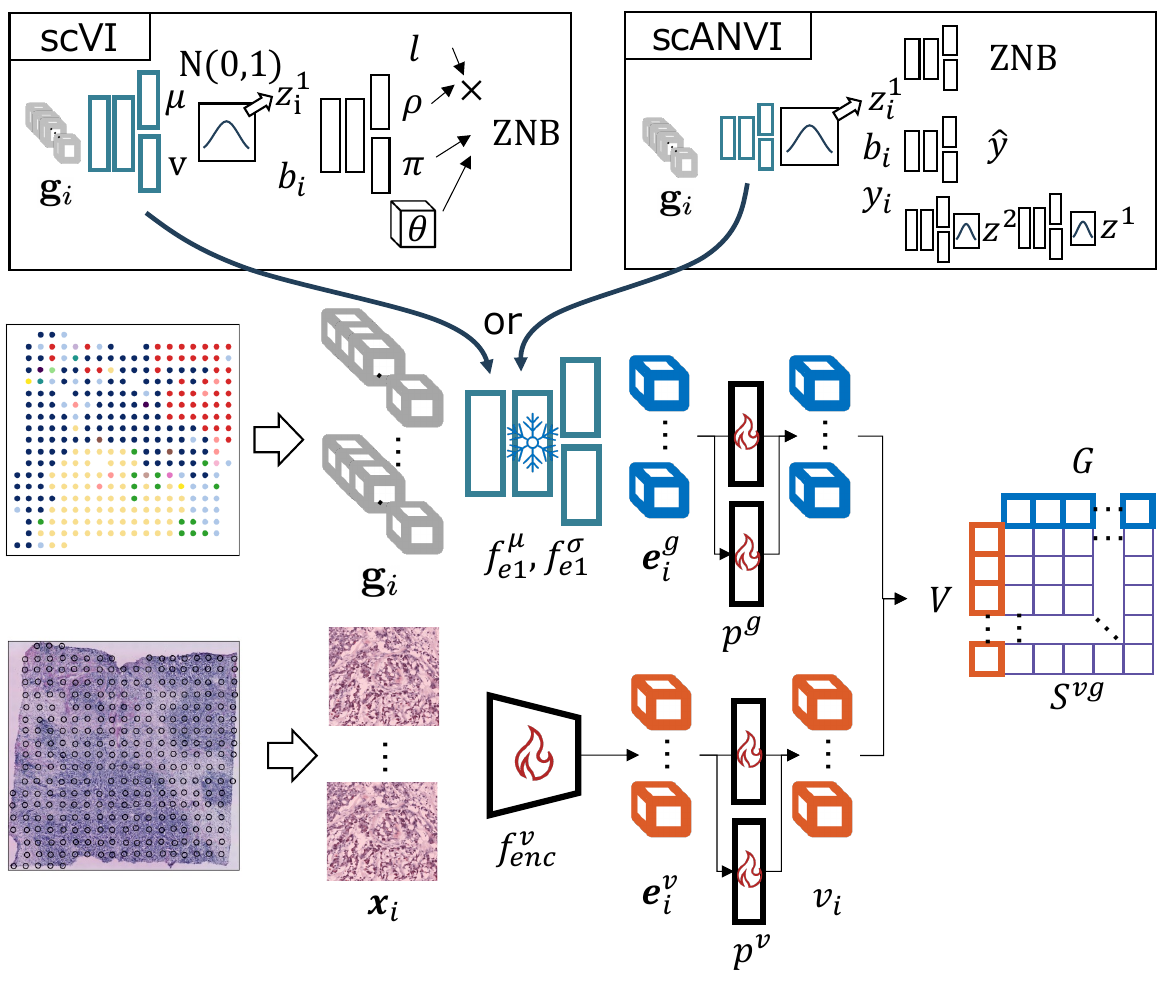}
    \caption{Overview of our method. scVI or scANVI are trained with the gene expression, and the encoder is reused for contrastive learning. }
    \label{fig:overview}
\end{figure}

Fig. \ref{fig:overview} shows an overview of our representation learning framework. 
Our framework consists of two steps.
In the first step, we obtain a robust gene expression encoder for batch effect.
In this step, we introduce variational inference for batch effect correction \cite{lopez2018deep}.
Second, we use the encoder in the first step to perform contrastive learning for image recognition with paired images and gene expression. 

\subsection{Training of batch-agnostic gene encoder}
To obtain the batch-agnostic gene encoder, we use a variational inference technique (scVI \cite{lopez2018deep}) widely used for batch correction of single-cell analysis.
While the encoder ideally disregards patient-specific information to ensure the encoded space aligns with a normal distribution, in a standard VAE, patient information is necessary for reconstruction. To address the problem, scVI delegates the reconstruction of patient-specific information (i.e., using patient conditioning for the decoder) and achieves batch agnostic encoder while preserving the natural feature expression of genes.


We use the pairs of a gene expression $\mathbf{g}_i \in \mathbb{R}^{N_g}$ and its patient label $\{\mathbf{g}_i, b_i\}^{N_s}_{i}$ for the training of scVI, where $N_g$ is number of genes, $i$ indicates $i$-th sample, and $N_s$ is the number of total samples. Here, we assume a latent variable $\mathbf{z}^1_i \sim \mathrm{N}(\mathbf{0}, \mathbf{1})$ and the observation model as a zero-inflated negative binomial distribution $\mathbf{g}_i \sim \mathrm{ZNB}(\bm{l}_i \bm{\rho}_i, \bm{\theta}_k, \bm{\pi}_{ik})$, where $\bm{l}_i$ is library size for gene, $\bm{\rho}_i$ is normalized gene expression, $\bm{\theta}$ is inverse dispersion, and $\bm{\pi}$ is zero inflated probability.

In the encoding process, we estimate the parameters $\bm{\mu}_{z^1_i}$ and $\bm{\sigma}_{z^1_i}$ for the variational distribution of $\mathbf{z}^1_i$ with neural networks: $\bm{\mu}_{z^1_i}=f_{e1}^{\mu}(\mathbf{g}_i)$ and $\bm{\sigma}_{z^1_i}=f_{e1}^{\sigma}(\mathbf{g}_i)$. 
Then, $\mathbf{z}^1_i$ is obtained with reparametrization tricks from $\mathrm{N}(\bm{\mu}_{z^1_i}, \bm{\sigma}_{z^1_i})$. 
In the decoding process, we use patient labels for condition and estimate the parameters for $\mathrm{ZNB}$ from $\mathbf{z}^1_i$ and $b_i$. The $\bm{\rho}_i = f_d^{\rho}(\mathbf{z}^1_i, b_i), \bm{\pi}_{ig} = f_d^{\pi}(\mathbf{z}^1_i, b_i)$ are obtained with two neural networks $f_d^{\pi}, f_d^{\rho}$, and $\bm{l}_i$ is obtained by summing the gene expression values of $\mathbf{g}_i$. 
We prepare a trainable vector for $\bm{\theta}_g$ and optimize $\bm{\theta}_g$ during training.

We maximize evidence lower bounds (ELBO) on the log evidence for training.
The ELBO has two terms: the first term is a log-likelihood between the observed distribution and estimated $\mathrm{ZNB}(\bm{l}_i \bm{\rho}_i, \bm{\theta}_g, \bm{\pi}_{ig})$, and the second term is Kullback-Leibler divergences between $\mathrm{N}(\bm{\mu}_{z^1_i}, \bm{\sigma}_{z^1_i})$ with $\mathrm{N}(\mathbf{0}, \mathbf{1})$.

In addition, since gene expression contains genes unrelated to the image, we also utilize subtype conditioning to obtain latent space close to the image. 
We train scANVI \cite{xu2021probabilistic} in a semi-supervised framework with few image annotations.

We use two sets for the training of scANVI: one is labeled data  $\{ \mathbf{g}_i, y_i, b_i \}$ and the other is unlabeled data $\{ \mathbf{g}_i, b_i \}$, where $y_i$ is an image annotation ({\it i.e.,} subtype label).
In addition to the VAE assumption, we assume a class-specific latent variable $z^2_i$, and the class labels follow a categorical distribution $y_i \sim \mathrm{Cat}(\pi_c)$ and also maximize ELBO for the training. A detailed explanation of the network architecture and training objectives is provided in \cite{xu2021probabilistic}.


\subsection{Details of contrastive learning with encoder}
As shown in Fig. \ref{fig:overview}, we update the image encoder $f^{v}$ and projectors $p^v$ and $p^g$ by reusing the trained gene encoder $f_{e1}^{\mu}$ and $f_{e1}^{\sigma}$ in a contrastive learning manner. The aim of this training is to get image representation, so we fixed the gene encoder.
Since the gene encoder can correct the batch effect and extract batch-agnostic features, we expected to obtain consistent signals from ST data by the encoder.


Given a batch of pairs of patch images $\mathbf{x}_i$ and gene expressions $\mathbf{g}_i$, the image and gene embeddings are extracted by the image encoder $f^{v}$ and the gene encoder $f_{e1}^{\mu}$ and $f_{e1}^{\sigma}$. The embedding $z^1_i$ is obtained by a reparametrization trick for the gene embedding. The both embeddings are converted by projectors $p^{v}$ and $ p^{g}$, and projected image embedding $\mathbf{e}^v_i = p^{v}(f^{v}(\mathbf{x}_i))$ and gene embedding $\mathbf{e}^g_i = p^g(z^1_i)$ are obtained. 

We explore the following losses for contrastive learning:
\begin{itemize}
    \item {\bf Symmetric InfoNCE \cite{oord2018representation} ($L_\mathrm{SI}$):} The $L_\mathrm{SI}$ is widely used for contrastive learning such as CLIP \cite{radford2021learning}. $L_\mathrm{SI}= - \frac{1}{2} (L_v + L_g)$, where $L_v = \frac{1}{B} \sum^B_i \log \frac{\exp(S^{vg}_{i, i})}{\sum_{j}^{B} \exp(S^{vg}_{i, j})}$, and $L_g = \frac{1}{B} \sum^B_j \log \frac{\exp(S^{vg}_{j, j})}{\sum_{i}^{B} \exp(S^{vg}_{i, j})} )$. Here, $S^{vg}_{i,j}$ indicates similarity of $\mathbf{e}^v_i$ and $\mathbf{e}^g_j$ and $B$ is mini batch size.
    \item {\bf Weighted Symmetric InfoNCE \cite{xie2024spatially} ($L_{\mathrm{WSI}}$):} Since nearby spots in the same whole slide image have similar trend, weights have been introduced to reflect the similarity. The weight is calculated from pair-wise similarity of images $S^{vv}$ and genes $S^{gg}$: $w = \mathrm{softmax_j(\frac{(S^{vv}_{i, j} + S^{gg}_{i, j})}{2\tau} )}$, where $\tau$ is temperature and $w \in \mathbb{R}^{B\times B}$. Then, the loss is calculated as follows: $L_v = \frac{1}{B} \sum^B_i w^v_{i,i} \log \frac{\exp(S^{vg}_{i, i})}{\sum_{j}^{B} \exp(S^{vg}_{i, j})}$.  $L_g$ is calculated in the same manner.
    \item {\bf Separated Weighted Symmetric InfoNCE ($L_{\mathrm{SWSI}}$)}: Since the same weights are used for $L_v$ and $L_g$, the similarity of each modality may be obscured. To reflect the trends of both modalities, we also compare a separate weighting approach, where $L_v$ and $L_g$ are weighted individually: $w^v = \mathrm{softmax}_j(S^{vv}_{i, j})$ and $w^g = \mathrm{softmax_j}(S^{gg}_{i, j})$.
    
\end{itemize}

\section{Experimental results}

\noindent
{\bf Dataset.}
In our experiments, we used a publicly available dataset of breast cancer with HER2 positive \cite{andersson2020spatial}, which consists of 36 slides captured from 8 patients with a $\times$20 magnification. Since the feature of one patient showed a markedly different trend compared to the others, we used 30 slides from 7 patients. 
Gene expression was captured on spots with a diameter of 100 $\mu m$, and the spots were spaced 200 $\mu m$ apart from each other.
We cropped the images to a size of $224 \times 224$ pixels from the center of the spot, pairing them with corresponding gene expression. Totally, $11,555$ image-gene expression pairs are obtained. For each patient, one WSI was annotated with six subtype labels, covering $2,816$ annotated samples. The class labels include invasive cancer, connective tissue, immune infiltrate, breastlands, cancer in situ, and adipose tissue.
We employed patient-level leave-one-out, and the performance of each method was evaluated using the mean of ten iterations for each fold.

\noindent
{\bf Implementation Details.}
We employed the ResNet50 model \cite{he2016deep} pre-trained on Imagenet \cite{russakovsky2015imagenet} for the image encoder. We used the implementation of the scVI toolkit for scVI and scANVI  \cite{lopez2018deep}. For the subtype annotation of scANVI, we used subtype labels other than test. These models were configured with a latent dimensionality of 30 and two layers. 
For the projector of image and gene, we used the same implementation of \cite{xie2024spatially}.
Our model was trained using one NVIDIA A6000 GPU with the AdamW optimizer, a batch size of 1024, and a learning rate of 0.001 for five epochs. 

\subsection{Evaluation}
To evaluate the effectiveness of our contrastive learning method, we assess a classification task as a downstream task after representation learning (pre-training) using our approach. We compared our method with one baseline method and two contrastive methods: 1) ResNet50, which was pre-trained on Imagenet; 2) CLIP \cite{radford2021learning}; and 3) BLEEP \cite{xie2024spatially}, which are common contrastive learning methods for images and gene expressions. 
These methods, except ResNet50, are used for pre-training, and the pre-trained models are subsequently fine-tuned on the label data.


\begin{table}[t]
    \centering
    \begin{tabular}{cc ccc} \hline
        Method & Loss & Enc. & Acc. & macro-F1 \\ \hline
        ResNet50 \cite{he2016deep} & - & - & \bf{0.677} & \underline{0.343} \\  \hdashline
        CLIP \cite{radford2021learning} & $L_\mathrm{SI}$ & - & 0.420 & 0.217 \\
        BLEEP \cite{xie2024spatially} & $L_\mathrm{WSI}$ & - & 0.507 & 0.084\\ \hdashline
        \multirow{3}{*}{Ours} & $L_\mathrm{SI}$ & \multirow{3}{*}{scVI} & 0.583 & 0.322\\
         & $L_\mathrm{WSI}$ &  & 0.593 & 0.180 \\ 
         & $L_\mathrm{SWSI}$ &  & 0.588 & 0.155 \\ 
        \hdashline
        \multirow{3}{*}{Ours}  & $L_\mathrm{SI}$ & \multirow{3}{*}{scANVI} & 0.601 & \bf{0.354} \\
         & $L_\mathrm{WSI}$ &  & 0.599 & 0.203 \\
         & $L_\mathrm{SWSI}$ &  & \underline{0.622} & 0.266 \\ 
        \hline
    \end{tabular}
    \caption{Comparison with previous contrastive learning methods.}
    \label{tab:my_label}
\end{table}

\begin{figure}
    \centering
    \includegraphics[width=\linewidth]{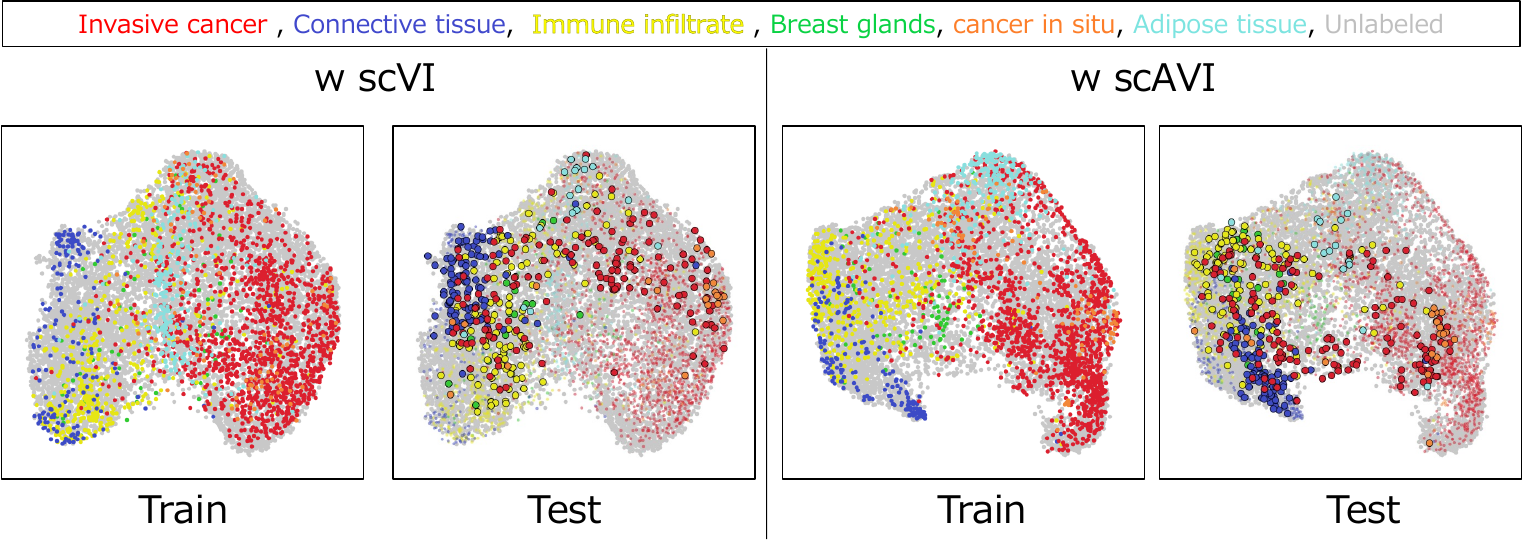}
    \caption{Visualization of feature space of our method with scVI and scANVI with $L_\mathrm{SI}$ colorized by the subtype label.}
    \label{fig:batch}
\end{figure}

Table \ref{tab:my_label} shows the result of comparisons.
Loss and Enc. Indicate the loss function and gene encoder used for training our method.
Since the scale of our dataset is smaller than Imagenet, our method is inferior to Imagenet pre-trained ResNet50.

\noindent
{\bf Effect of the batch-agnostic encoder.}
Our batch-agnostic method outperforms CLIP and BLEEP.
This suggests that our batch-agnostic contrastive learning can extract features for image recognition than previous methods because the gene encoder can extract batch-agnostic features from multiple patients, and it provides a consistent signal from ST.


\noindent
{\bf Effect of image conditioning.}
Our method using the scANVI encoder was slightly outperformed using the scVI encoder
The difference in performance between using scVI and scANVI was not distinct despite using class labels directly related to the target task.

To explore the reason for the small improvement, we examined the feature spaces of our method of both encoders. 
Fig. \ref{fig:batch} shows the feature space of our framework with the encoder of scVI or scANVI for training and test data.
In the training data, the representation of scANVI shows a higher consistency with class labels, especially for \textcolor{green}{breast glands}. However, in the test data, the consistency with the class labels was not significantly different between the embeddings using scVI and scANVI.

The gene representation with scANVI was forcibly pushed into class condition for the image, which is unrelated to gene expression. 
Since the image class labels are determined by the human inspection of images, the class definitions may not be fit for gene expression profiles, potentially disrupting the natural representation of gene expression. Achieving more natural representations may require conditioning classes aligned with gene expression and carefully selecting input genes that are related to morphology. 

\noindent
{\bf Comparisons of Contrastive loss.}
Overall, $L_\mathrm{SI}$ achieves the best performance.
The $L_\mathrm{WSI}$ and $L_\mathrm{SWSI}$ can account for sample similarities, but the weights are highly influenced by temperature. 
Moreover, in our framework, only the gene distribution is regularized to follow a normal distribution $N(0, 1)$. The gene and image distributions do not share similar trends, which makes utilizing both distributions difficult.

\section{Conclustion}
\label{sec:Conclustion}

We proposed a batch-agnostic contrastive learning framework with gene expression of spatial transcriptomic (ST) and image for subtype recognition of image. 
Gene expression of ST contains extensive information about cellular activity in addition to image features. However, it is challenging to extract consistent signals from ST due to the batch effect, which is a gap caused by experimental batches. To extract consistent signals from ST for image recognition, we utilized a batch-agnostic encoder derived from a variational inference-based training schema.
Experiments demonstrated the effectiveness of the proposed method with the publicly available dataset. Our method outperforms the previous method. In addition, we confirmed that only using contrastive learning for the training is not effective in our dataset.

\section{Limitations}
In our experiments, we used a widely used dataset that contains a relatively large number of patients with the same condition. Although the scale of the dataset is relatively larger than other datasets, the dataset scale is still small for contrastive learning.
More data is required to outperform the model-trained large-scale dataset such as Imagenet.
Although challenging due to the capturing cost, larger paired datasets of image and ST are required to further explore the potential of contrastive learning with image and spatial transcriptomic. While large-scale datasets such as hest-1k \cite{jaume2024hest} have been published, the datasets contain samples from multiple organs with various diseases, and the samples have a big data gap more than the patient gap that we addressed.
Addressing the gap problem is unavoidable when leveraging ST.
This study is the first attempt to treat the data gap problem of ST, and the attempt is important for pathological images with multi-modal data.

\section{compliance with ethical standards}
This research study was conducted retrospectively using human subject data made available in open access by Andersson {\it et al.} \cite{andersson2020spatial}. 
Ethical approval was not required, as confirmed by the license attached with the open-access data.

\section{Acknowledgements}
This work was supported by JSPS KAKEN JP24KJ2205 and JP23K18509, JST ACT-X, Grant Number JPMJAX21AK, Japan Agency for Medical Research and Development (AMED) grant 24ama221609h0001(P-PROMOTE) (to YK), National Cancer Center Research and Development Fund 2024-A-6 (to YK).
\bibliographystyle{IEEEbib}
\bibliography{refs}

\end{document}